# Ensemble2: Anomaly Detection via EVT--Ensemble Framework for Seasonal KPIs in Communication Network

SHIYANG WANG*

School of Information and Software Engineering, University of Electronic Science and Technology of China, Chengdu 610054, China

KPI anomaly detection is one important function of network management system. Traditional methods either require prior knowledge or manually set thresholds. To overcome these shortcomings, we propose the Ensemble[2] framework, which applies ensemble learning to improve exogenous capabilities. Meanwhile, automatically adjusts thresholds based on extreme value theory. The model is tested on production datasets to verify its effectiveness. We further optimize the model using online learning, and finally running at a speed of ~10 pts/s on an Intel i5 platform.

**CCS CONCEPTS** • Networks~Network services~Network management • Computing methodologies~Machine learning~Learning paradigms~Unsupervised learning~Anomaly detection • Mathematics of computing~Probability and statistics~Statistical paradigms~Time series analysis

## 1 INTRODUCTION

In communication network, anomaly detection has long been a research hot spot. Communication network is a complex system, which can be defined by network element model and management model. The network element model states that each node in a communication network represents one base station. A base station receives signals via antennas, and transmits them through underground optical fiber sub-network. Meanwhile, the dedicated server within each station collects data. And the network management system performs analysis on these data, such as anomaly detection, root cause analysis, or geolocalization.

Anomaly detection is one of the many functions of the network management system, dealing with time series data collected throughout network. These time series describe all aspects of the running status of the network, in the form of key performance indicators (KPI), including call drop rate, CPU utilization rate, and failed interface calls. These indicators reflect the health status of the network. Therefore, closely monitoring these indicators both ensures the quality of services and aids in resolving customer complaints. For example, in remote areas where base stations are sparsely distributed, during the period of international conference, mobile base stations can be deployed to improve service quality. Besides, frequent interface failure may indicate a compatibility issue. Moreover, in the 5G era, the Network Data Analysis Function (NWDAF) is specifically aimed at quality assurance, via AI and ML technologies. In short, anomaly detection is an essential component of network management system.

* Corresponding Author:Shiyang Wang, Email: shiyangw0118@163.com

For time series anomaly detection, SOA solutions include threshold-based approaches, machine learning algorithms, and deep models. These traditional methods either complex in tuning, or require prior knowledge or human interference. For example, neural network is one of the over-parameterized models, the scale of parameter is often much larger than that of training data. Thus, it requires to validate whether the model learned from data, or simply remembered its training set. On the other side, threshold-based methods require prior knowledge and fine-tuning its parameters.

To overcome these shortcomings, we propose Ensemble[2], an ensemble framework enhanced by extreme value theory. Base learners are chosen from classic machine learning algorithms, since they are simple and interpretable. These methods include ARIMA, STL, and SVR. The framework involves voting on these base learners, to improves accuracy, robustness, and extrapolation performance. Extreme value theory, first proposed in 1950s, is often used to model rare events with a heavy-tailed distribution and achieves competing results. Therefore, the threshold is automatically adjusted using Generalized Pareto Distribution. The point is, our approach makes no prior assumption on data distribution.

Our main contributions include:

1. We propose and implement Ensemble[2], a periodic KPI anomaly detection framework based on ensemble learning and extreme value theory, achieving good result in different types of data with heterogenous base learners.
2. Lilelihood moment estimation is used to calculate parameters of Generalized Pareto Distribution, and permutation entropy to interpret prediction results.
3. Batch processing is further optimized into its streaming counterpart. The source code is published online (https://github.com/Shi-YangWang/ensemble-square-online). And finally it achieves a running speed of ~10 pts/s on an Intel i5 platform.

## 2 ENSEMBLE[2]

### 2.1 ARIMA

ARIMA itself consists of several other models, including two smoothing filters AR and MA, plus an r-order differentiation. By integrating differentiation, ARIMA can be applied on non-stationary time series. The model can be expressed as:

$$(1 - \phi_1 B - \cdots - \phi_p B^p) \cdot (1 - B)^d y_t = c + (1 + \theta_1 B + \cdots + \theta_q B^q) \varepsilon_t$$

, where B represents the back shift operator:

$$By_t = y_{t-1}$$

For time series, stationarity refers to the characteristic that the variance of sample data does not change over time. Intuitively, the more high-frequency components, the higher order of differentiation is required to stablize it.

The advantages of ARIMA are obvious. It is fast and simple, and the outputs roughly describe the outline of data. However, shortcomings are also straightforward. For different KPI indicators, parameters need to be recalculated. This increases the complexity in time. Meanwhile, some time series appear to have a certain degree of periodicity. Due to the existence of high-frequency components, valuable information will be shadowed by irregular noise.

## 2.2 STL

STL (Seasonal Trend decomposition using Loess) is a decomposition method for time series, and the implementation details are are follows:

1. Given a window of size m, the trend component of input time series $\hat{T}_t$ is obtained by applying smoothing filter. When m is even, the smoothing filter is applied twice. Odd, once.

2. Subtract trend component from original data, and a remaining terms $y_t - \hat{T}_t$ is obtained.

3. Reshape the remaining term into n periods, and the length of each period is constant, $C$. Then, average on these periods to obtain seasonal component:

$$\widehat{S}_t = \sum_{i=0}^{N} y_{t_0+Ci} - \hat{T}_{t_0+Ci}$$

4. Subtracting trend and seasonal components from the original time series, and the residual term is obtained:

$$\hat{R}_t = y_t - \hat{T}_t - \hat{S}_t$$

STL integrates Loess regression to improve robustness. Loess, also called Savitzky-Golay filter, is a regression algorithm that uses k nearest neighbors to perform least squares regression. The implementation details are listed below:

1. For each timestamp $y_t$, take k nearest neighbors $y_i$ ($0 \le i < k$), where k is parameter. Calculate the normalized distance: $\Delta = \dfrac{d - \min}{\max - \min}$, where max represents the maximum distance among them while min the minimum.

2. Map distance $\Delta$ to weight $w$ through tri-cubic function:

$$w = f(\Delta) = (1 - \Delta^3)^3$$

3. Calculate smoothed value $\hat{y}_t$ using least squares regression:

$$Y = X\theta$$

$$\hat{\theta} = (X^T X)^{-1} X^T Y$$

$$\hat{y}_t = x\hat{\theta}$$

It can be seen that the introduction of k nearest neighbors makes the model robust to outliers as a whole.

## 2.3    LS-TSVR [1]

The regression function of least squares twin support vector regression can be obtained by solving the following QPPs:

Minimize

$$\left\{ \frac{1}{2} \left\| Y - \varepsilon_1 e - (K(X, X^T)\omega_1 + b_1 e) \right\|^2 + \frac{C_1}{2} \xi^T \xi \right\}$$

*s.t.*

$$Y - (K(X, X^T)\omega_1 + b_1 e) = \varepsilon_1 e - \xi$$

Minimize

$$\left\{ \frac{1}{2} \left\| Y - \varepsilon_2 e - (K(X, X^T)\omega_2 + b_2 e) \right\|^2 + \frac{C_2}{2} \varsigma^T \varsigma \right\}$$

*s.t.*

$$(K(X, X^T)\omega_2 + b_2 e) - Y = \varepsilon_2 e - \varsigma$$

, where $K(X, X^T)$ is the kernel matrix after the mapping of empirical feature space.

Substitute the equality constraint in equation formula (8) into the objective function to obtain the following formula:

$$L(\omega_1, b_1, \xi) = \frac{1}{2} \left\| Y - \varepsilon_1 e - (K(X, X^T)\omega_1 + b_1 e) \right\|^2 + \frac{C_1}{2} \left\| K(X, X^T)\omega_1 + b_1 e + \varepsilon_1 e - Y \right\|^2$$

Setting the gradient of formula (6) with respect to $\omega_1$ and $b_1$ to zero, we can obtain

$$\frac{\partial L(\omega_1, b_1, \xi)}{\partial \omega_1} = -K(X, X^T)^T (Y - K(X, X^T)\omega_1 - b_1 e - \varepsilon_1 e) + K(X, X^T)^T C_1 (K(X, X^T)\omega_1 + b_1 e + \varepsilon_1 e - Y) = 0$$

$$\frac{\partial L(\omega_1, b_1, \xi)}{\partial b_1} = -e^T (Y - K(X, X^T)\omega_1 - b_1 e - \varepsilon_1 e) + e^T C_1 (K(X, X^T)\omega_1 + b_1 e + \varepsilon_1 e - Y) = 0$$

After spatial mapping, the nonlinear solution process is consistent with the linear solution process, $\begin{bmatrix} \omega_1 & b_1 \end{bmatrix}^T$ can be obtained as follows:

$$\begin{bmatrix} \omega_1 \\ b_1 \end{bmatrix} = \left[ \begin{bmatrix} K(X,X^T)^T \\ e^T \end{bmatrix} \begin{bmatrix} K(X,X^T) & e \end{bmatrix} + \frac{1}{C_1} \begin{bmatrix} K(X,X^T)^T \\ e^T \end{bmatrix} \begin{bmatrix} K(X,X^T) & e \end{bmatrix} \right]^{-1} (Y - e\varepsilon_1) \left( 1 + \frac{1}{C_1} \right)$$

Using the similar solving method, $\begin{bmatrix} \omega_2 & b_2 \end{bmatrix}^T$ can be obtained as follows:

$$\begin{bmatrix} \omega_2 \\ b_2 \end{bmatrix} = \left[ \begin{bmatrix} K(X,X^T)^T \\ e^T \end{bmatrix} \begin{bmatrix} K(X,X^T) & e \end{bmatrix} + \frac{1}{C_2} \begin{bmatrix} K(X,X^T)^T \\ e^T \end{bmatrix} \begin{bmatrix} K(X,X^T) & e \end{bmatrix} \right]^{-1} (Y - e\varepsilon_2) \left( 1 + \frac{1}{C_2} \right)$$

Finally, the regression function of least squares twin support vector regression can be obtained as follows:

$$f(x) = \frac{1}{2} K(x,X^T)(\omega_1 + \omega_2) + \frac{1}{2}(b_1 + b_2)$$

It can be seen that the solution of LS-TSVR naturally contains the least squares solution of pseudo inverse [2], that is, $G^+ = (G^T G)^{-1} G^T$ . We know that the generalized inverse matrix $X$ needs to satisfy several equations in the following conditions:

1. $AXA = A$
2. $XAX = X$
3. $(XA)^* = XA$
4. $(AX)^* = AX$

, where $A^*$ stands for conjugate transpose of matrix $A$ .

The least squares solution $G^+$ of the inverse matrix is the $\{1,2\}$ - inverse of G. Therefore, we can consider using Moore-Penrose generalized inverse [3] instead of pseudo inverse. Since the Moore-Penrose generalized inverse is $\{1,2,3,4\}$ - inverse , better properties can be obtained. In addition to that, Moore-Penrose generalized inverses can be computed quickly using SVD-based algorithms. The Moore-Penrose generalized inverse also naturally supports online learning, through inversion of block matrix, as is implemented in the breadth learning system.

### 2.4    Ensemble Learning

Ensemble accomplishes a task by building and combining multiple learners. Common techniques include Bagging, Boosting, and Stacking. Bagging samples from training sets, to obtain multiple models. And combine them by voting or averaging, to get the final output. Boosting goes another way. It tries to make up for the errors produced by previous model, and iterate till convergence. Empirically, the more iterations, the less error. Stacking involves several different types of base models, and combine them by training a meta learner.

## 2.5 Extreme Value Theory

First proposed in 1950s, extreme value theory aims to model rare natural disasters such as hurricanes, floods, and earthquakes.

Let's take the specific problem of modeling floods as an example. The water level is collected at a fixed interval, and a time series of water level can be obtained. Predictably, this time series is periodic, and the noise can be modeled appropriately using Gaussian. But, the question is, how to model outliers in it? We can reshape the time series into several periods. Set a threshold and model the probability distribution for extreme values in each time period. This method of modeling extreme events is called a peak over threshold (POT) approach.

The distribution of these extreme values is different from classical distributions, such as binomial, geometric, or Gaussian. With the research progressing, Gumbel, Weibull, and Pareto distribution have been proposed one after another. They consist a new family of heavy-tailed distributions.

After meticulous consideration, Generalized Pareto Distribution (GPD) [4, 5] is chosen to model the outliers in KPI. Cumulative distribution function of two-parameter GPD is:

$$F(x;\sigma,k) = \begin{cases} 1 - (1 - \dfrac{kx}{\sigma})^{1/k}, k \neq 0 \\ 1 - e^{-x/\sigma}, k = 0 \end{cases}$$

, where $\sigma$ controls the scale while $k$ controls the shape.

The quantile function is:

$$z_q \simeq t + \frac{\sigma}{k}((\frac{qn}{N_t}) - 1)$$

, where $t$ is a threshold, $q$ is the desired probability, $n$ is the total number of observations, and $N_t$ the number of peaks.

## 2.6 Ensemble²

Flow chart of the Ensemble² framework is shown below:

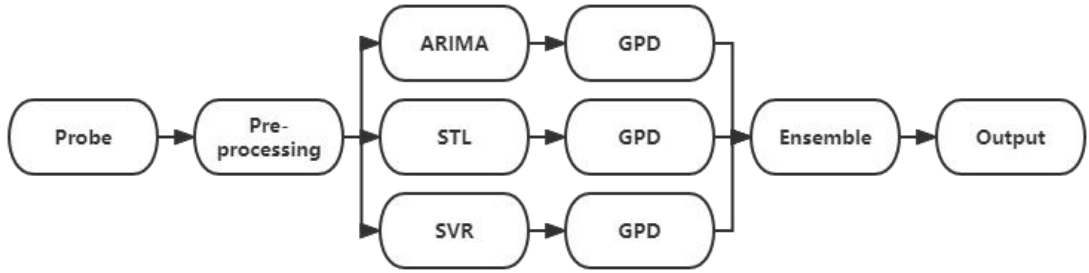

Figure 1: Schematic diagram of Ensemble[2] framework

The anomaly detection algorithm is described as:

---

ALGORITHM 1: Anomaly detection method

---

Initialize A = $\phi$ , C = $\phi$

Initialize $\varepsilon_1$ = quantile( $X_{train}$ , q)

Initialize $\varepsilon_2$ = quantile( $X_{train}$ , $\theta$ )

for k > 0 do

  if $e_k > \varepsilon_1$

    add to anomaly set A

  else if $e_k > \varepsilon_2$

    add to candidate set C

    re-estimate $\varepsilon_1$ and $\varepsilon_2$ using GPD quantile function

  else

    k = k + 1

  end if

end for

---

## 3 RELATED WORK

The existing solutions to time series anomaly detection are mainly classified into three categories: unsupervised algorithms, deep models, and cutting-edge methods.

Unsupervised algorithms aim to smooth the time series, and calculate its difference from the original signal. Moving average [6] uses a direct approach, replacing the centermost value by an average of the other points in the window. The problem is when a spike enters the window, the smoothed curve is abruptly distorted. Thus weighted moving average is used to tackle this problem, which gives less weight to edge points. Exponential smoothing is another solution. It works by mixing the result from previous time step together with data from current step. And the degree of mixing can be controlled by parameters. Though defined recursively, it can be reformatted into a Taylor expansion of exponential function. And there it gets its name. Matrix profile [7] is another unsupervised algorithm. It is officially a data structure containing nearest information for all subsequences of a given length in time series. Matrix profile is a panacea feature in data engineering.

Deep models are known for their ability to capture long term dependencies. Zhao et al [8] integrate sparsely connected neural networks into autoencoder framework and improve its robustness. Liu el al [9] goes further in the idea of ensemble. By selectively removing existing and forming new connections, different network structures are searched over time to find the most appropriate. TCN [10] is a combination of technologies related to convolution, including causal convolution, dilated convolution, and shortcut connection. Causal convolution is proposed to solve the problem of feature leakage, i.e. the output at a later point can only be affected by previous observations. Residual connection is a trick first proposed in ResNet to form identity mapping.

Cutting-edge methods are yet another approach in anomaly detection. Du et al [11] closely follows advances in domain adaptation, and propose a temporal covariate shift (TSC) problem. Research team further solves this problem with greedy algorithm. Moreira et al [12] uses visibility map to translate time series into a complex network, and prediction is based on node similarity. HTM [13] adopts the idea of breadth learning to detect outliers. The encoder obtains a sparse distributed representation (SDR) of input. Then the sequential memory optimizes its structure through a recursive procedure. Finally anomaly score is calculated by counting active cells in each cortical column.

## A    APPENDICES

## A.1    Experiment

### A.1.1    Pre-processing

We validate the proposed algorithm using production data in communication network. The datasets contain a total of 26 KPI sequences, and each sequence is split into training set of 45 days and test set of 15 days. The sampling frequency varies from 1 to 5 minutes.

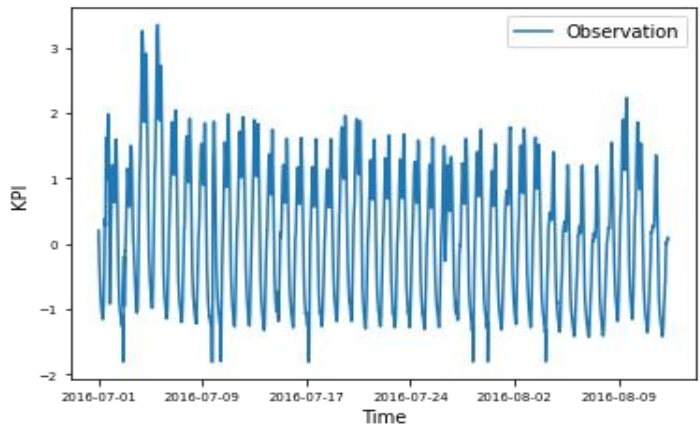

Figure 2: Overview of raw datasets

In pre-processing, we first perform data probing on the datasets. From the line chart, we can roughly see that the data is distributed periodically, with a period of around one day (calculated at a sampling frequency of 1 minute, 1 day contains 1440 sample points). We used pandas to calculate the maximum, minimum, quantile and other characteristics of the series, as shown in the following table:

Table 1: KPI data statistics

| Count | Avg | Std | Min |
|-------|-----|-----|-----|
| 10960 | 0.155 | 1.051 | l-1.809 |

| 25% | 50% | 75% | Max |
|---|---|---|---|
| -0.869 | 0.137 | 0.137 | l3.348 |

Among these features, the maximum value is positive, the minimum value is negative, and the mean value is approximately zero, which is similar to a sine (cosine) function. Although the original datasets does not disclose the specific meaning of these KPIs for confidentiality reasons, we can speculate that it has been manufactured.

After data probing, the data needs to be normalized. Normalization is an essential step in deep models, for nonlinear functions like hyperbolic tangent are sensitive to data in a specific interval (such as [-1, 1]), and unaligned raw data will reduce learning efficiency. Although our proposed solution does not include a deep model, normalization is retained. The purpose is to obtain a generalized model.

A common method is zero-mean normalization:

$$x' = \frac{x - \mu}{\sigma}$$

, where $\mu$ is the sample mean and $\sigma$ is the sample variance. Obviously, the normalized data has a mean equal to 0 and a variance equal to 1.

### A.1.2 LS-TSVR

To verify the effectiveness of the optimized LS-TSVR algorithm, a test was performed on three days of data, with a total of 4320 sample points.

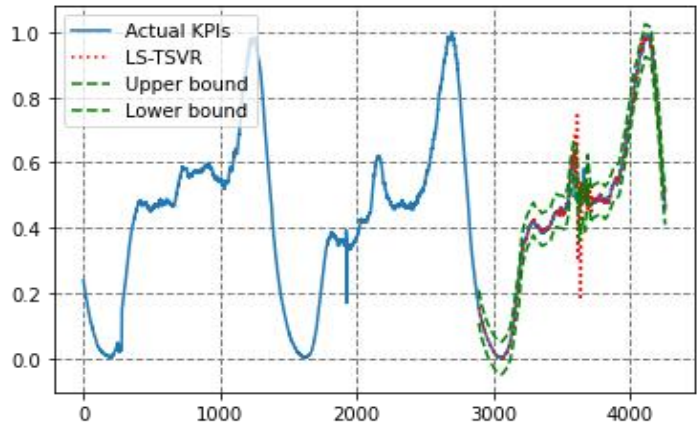

Figure 3: Fitness curve using LS-TSVR

Because the framework employ a reconstruction error-based method for anomaly detection, the purpose of this test is to verify the accuracy of the LS-TSVR algorithm for single-step prediction. Experimental result shows that the optimized LS-TSVR algorithm further lower the prediction error and reduces time consumption to a certain extent. This improvement comes from using the Moore-Penrose inverse instead of the least square inverse.

Table 2: Comparison result between SVR and LS-TSVR

| Model | SVR | LS-TSVR |
|-------|-----|---------|
| MSE | 0.003 | **0.002** |
| MAE | 0.051 | **0.014** |
| CPU time (s) | 224.684 | **146.119** |

It should be noted that LS-TSVR algorithm can perform interval prediction, i.e. predicting upper and lower bounds. Although range prediction is not used in the Ensemble2 framework, it is a potentially valuable feature.

### A.1.3  Ablation Experiment

We selected six representative chunks from the original datasets. Among them, segment 1 and 2 contain fewer high-frequency components, while segment 3 and 4 contain more high-frequency components. Segment 5 and 6 exhibits "arrhythmia" phenomenon.

Table 3: 6 chunks selected from raw datasets

| Chunk | KPI ID | Start Date | End Date |
|-------|--------|------------|----------|
| 1 | 9ee5879409dccef9 | May 1, 2017 | May 8, 2017 |
| 2 | 88cf3a776ba00e7c | Apr 24, 2017 | May 1, 2017 |
| 3 | affb01ca2b4f0b45 | May 8, 2017 | May 15, 2017 |
| 4 | 8bef9af9a922e0b3 | May 13, 2017 | May 20, 2017 |
| 5 | 046ec29ddf80d62e | Jul 25, 2016 | Aug 1, 2016 |
| 6 | 76f4550c43334374 | Oct 29, 2016 | Nov 5, 2016 |

On these six fragments, we performed ablation experiments. Among them, the precision rate, recall rate, and F1 value are calculated:

$$precision = \frac{TP}{TP + FP}$$

$$recall = \frac{TP}{TP + FN}$$

$$F - Score = \frac{2 \cdot precison \cdot recall}{precision + recall}$$

, where TP, FP, and FN are abbreviations for true positive, false positive, and false negative, respectively.

In the evaluation phase, we use a modified calculation method of F1 score: if the predicted anomaly point lies within a T step window from ground truth, it is considered as a match.

Table 4: Precision, recall, and F1 score of 6 chunks

| Chunk | E2nsemble | Without STL | Without ARIMA | Without LS-TSVR |
|-------|-----------|-------------|---------------|-----------------|
| | Precision | | | |
| | Recall | | | |
| | F1 score | | | |
| 1 | **0.935** | 0.609 | 0.763 | 0.898 |
| | 0.563 | 0.352 | 0.439 | 0.965 |
| | 0.703 | 0.446 | 0.557 | 0.930 |
| 2 | **0.881** | 0.696 | 0.652 | 0.686 |

| Chunk | E2nsemble | Without STL | Without ARIMA | Without LS-TSVR |
|---|---|---|---|---|
| | | Precision | | |
| | | Recall | | |
| | | F1 score | | |
| | 0.509 | 0.403 | 0.404 | 0.440 |
| | 0.645 | 0.510 | 0.499 | 0.536 |
| 3 | **1.000** | 1.000 | 1.000 | 1.000 |
| | 1.000 | 0.769 | 0.813 | 1.000 |
| | 1.000 | 0.870 | 0.897 | 1.000 |
| 4 | **1.000** | 1.000 | 0.857 | 1.000 |
| | 1.000 | 0.800 | 1.000 | 1.000 |
| | 1.000 | 0.889 | 0.923 | 1.000 |
| 5 | **0.938** | 0.725 | 0.938 | 0.688 |
| | 0.615 | 0.464 | 0.628 | 0.792 |
| | 0.743 | 0.566 | 0.752 | 0.736 |
| 6 | **0.928** | 0.675 | 0.928 | 0.542 |
| | 0.571 | 0.548 | 0.581 | 0.660 |
| | 0.707 | 0.605 | 0.715 | 0.595 |

Conclusions: (i) high-frequency noise in the data will reduce the accuracy of ARIMA, (ii) "arrhythmic" will reduce the accuracy of LS-TSVR, (iii) STL is not affected by "arrhythmia".

Possible Explanations: (i) the irregular high-frequency components will reduce the stationarity of the data, thus affecting ARIMA; (ii) LS-TSVR uses the least square approximation, bringing performance boost and sensitive to outliers; (iii) STL is robust to outliers due Loess regression.

Most importantly, Ensemble2 achieved the best accuracy in all experiments.

### A.1.4  Long Range Dependency

We use xgboost to evaluate importance of long-range dependencies. We set number of estimators to 10, proportion to 30%, maximum depth to 5, and learning rate to 0.1.

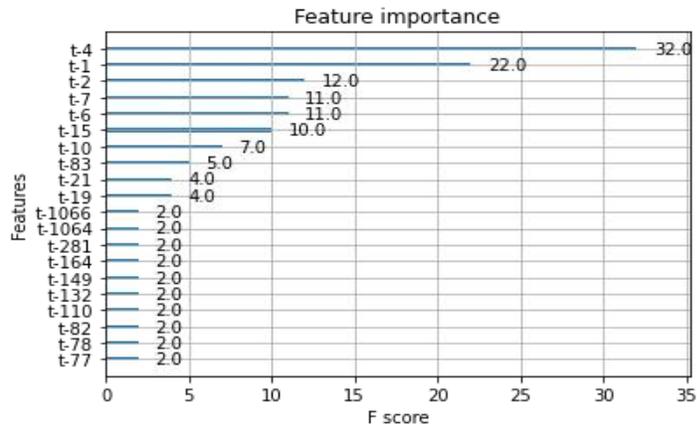

Figure 4: Ranking of time series characteristics at historical moments

As can be seen from the results, t-1 (historical data from 1 minute ago) and t-4 are the two most important features. The feature of t-2 is less than half as important as t-1. This shows that the data closer to the current time point is more important, and the distant historical data has little effect on the prediction accuracy. In other words, considering long-range dependencies will not increase accuracy, rather, it will increase the risk of overfitting. Therefore, the size of time window is set to 60 in the model, that is, the impact of historical data within 1 hour is considered.

### A.1.5   Permutation Entropy

Permutation entropy was proposed by Bandt and Pompe [14] to describe the complexity of chaotic dynamical systems. Because permutations themselves are sequences, they are also used to describe the complexity in time series. Compared to other metrics of (i.e. Matrix Profile), it has intrinsic physical meanings. The permutation entropy reflects the chaotic nature of time series, i.e. greater the permutation entropy, less information is contained, and vice versa.

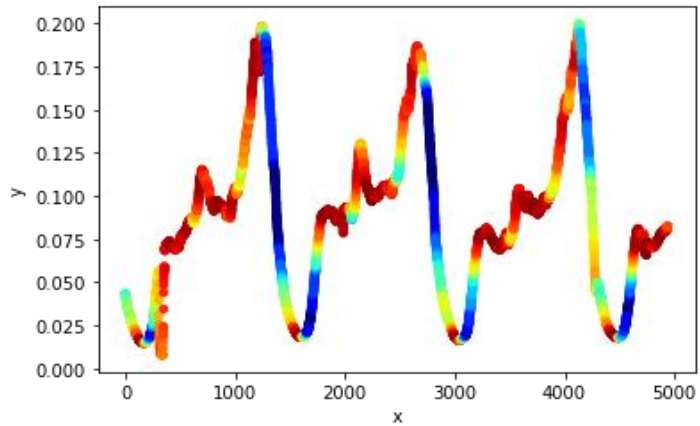

Figure 5: Permutation entropy plot of chunk1

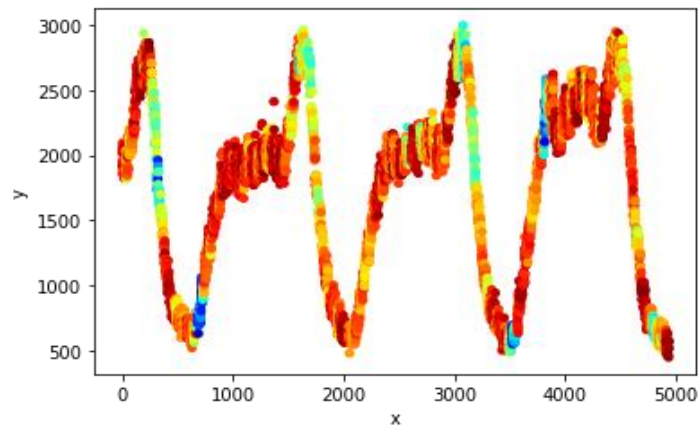

Figure 6: Permutation plot of chunk3

Take the time window size w=60, order=3, and the permutation entropy corresponding to chunk1 and chunk3 is shown above. In Figure 5 and 6, we map the permutation entropy to a color gradient and superimpose it on top of the original time series. Red color illustrates that the permutation entropy within this segment is larger and the sequence contains less information. On the contrary, blue indicates that the permutation entropy is relatively small, and the sequence contains more information. This is consistent with our intuition that chunk3 seems more chaos while chunk1 more regular.

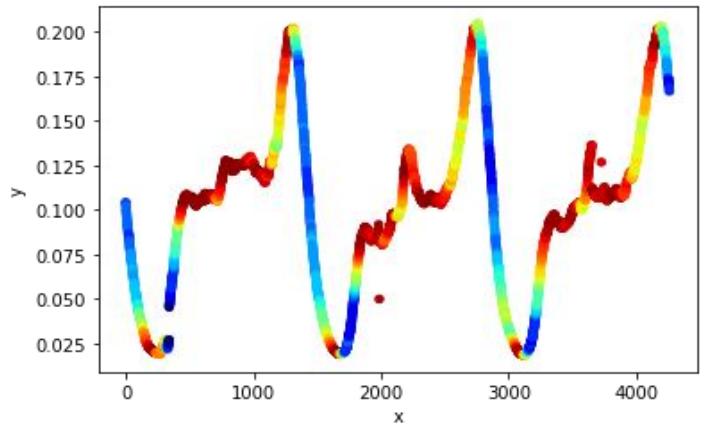

Figure 7: Permutation entropy analysis of LS-TSVR

Figure 7 corresponds to the time series used in Section *A.1.2*. Obviously, the LS-TSVR model gives fuzzy prediction results in chaos interval, and gives accurate prediction in regular interval.

### A.1.6 Optimization

We further replace base learners in our framework with online algorithms [15-17]. Online ARIMA cuts out MA filter, and quickly calculates the inverse matrix by Woodbury identity matrix. Online STL uses two alternating trend and seasonal filters to approximate Loess regression. Online SVR follows the idea of incremental SVR.:According to three-way decision theory, candidate set C is divided into support set S, error set E and residual set R. Online calculation is performed by incremental and decremental algorithms. In short, online algorithms relax some conditions in exchange for performance boosting. It is a trade-off.

Online algorithm was tested on synthetic datasets. It includes a total of 8*1440 sample points sampled on sine wave function, and anomaly is injected. To simulate a production environment, Gaussian noise was added to the datasets. On an Intel i5 platform, the program runs at a speed of ~10 pts/s.

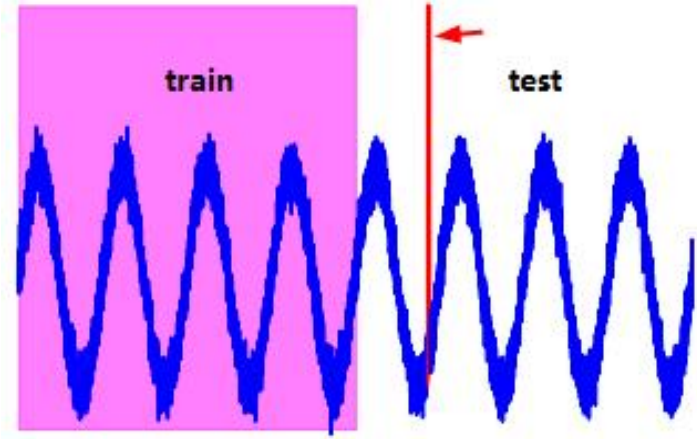

Figure 8: Anomaly detection using arbitrary datasets

It can be seem that, due to the absence of Loess regression in online STL, there was a tailing phenomenon. Fortunately, it can be absorbed by GPD and ensemble framework.

## A.2 Estimation of GPD Parameters

**Lemma [18].** *Let r.v.* $X \sim GPD(\sigma, k)$, *then* $E[(1 - bX)^r] = (1 + rk)^{-1}$.

**Proof.** $F(x) = 1 - (1 - bx)^{1/k}$, $b = k/\sigma$

$$f(x) = F'(x) = \frac{1}{\sigma}(1 - bx)^{1/k - 1}$$

$$E[(1 - bx)^r] = \int f(x)(1 - bx)^r \, dx = (1 + rk)^{-1} \quad \Box$$

**Corollary.** *(Moment Estimation)*

$$\hat{\sigma} = \frac{(\overline{x})^3}{2(s^2 + 1)}, \quad \hat{k} = \frac{(\overline{x})^2}{2(s^2 - 1)}$$

, *where* $\overline{x}$ *and* $s^2$ *denote sample mean and variance, respectively.*

**Corollary.** *(Likelihood Moment Estimation)*

*Let* $r = 0$, $\hat{k} = -\frac{1}{n}\sum(1 - bX_i)$

Let $r = -1$, $\frac{1}{n}\sum\frac{1}{1 - bX_i} - \frac{1}{1 - k} = 0$

| Authors' background / The form itself will not be published, Please Delete it on your final papers | | | |
|---|---|---|---|
| **Your Name** | **Title*** | **Research Field** | **Personal website** |
| WANG SHIYANG | Master Student | Anomaly Detection, Time Series Analysis, Network Management | |
| | | | |
| | | | |
| ***Title can be chosen from: master student, Phd candidate, assistant professor, lecture, senior lecture, associate professor, full professor*** | | | |